\documentclass[12pt]{article}

\usepackage{geometry}
\usepackage{tabularx}
\usepackage{amsmath, amssymb}
\usepackage{graphicx}
\usepackage{booktabs}
\usepackage{xcolor}
\usepackage{hyperref}
\usepackage{authblk}
\usepackage[numbers,sort&compress]{natbib}

\title{AyutthayaAlpha: A Thai-Latin Script Transliteration Transformer}

\author[1]{Davor Lauc}
\author[2]{Attapol Rutherford}
\author[3]{Weerin Wongwarawipatr}

\affil[1]{University of Zagreb, Faculty of Humanities and Social Sciences, Zagreb, Croatia\\
Email: \texttt{dlauc@ffzg.unizg.hr}}

\affil[2]{Chulalongkorn University, Department of Linguistics, Bangkok, Thailand\\
Email: \texttt{attapolrutherford@gmail.com}}

\affil[3]{Freyr Solutions, Helsinki, Finland\\
Email: \texttt{weerin.ann2019@gmail.com}}

\date{20 November 2024}

\begin{document}

    \maketitle

    \begin{abstract}
        This study introduces AyutthayaAlpha, an advanced transformer-based machine learning model designed for the
        transliteration of Thai proper names into Latin script. Our system achieves state-of-the-art performance with
        82.32\% first-token accuracy and 95.24\% first-three-token accuracy, while maintaining a low character error rate of
        0.0047.
        The complexity of Thai phonology, including tonal features and vowel length distinctions, presents
        significant challenges for accurate transliteration, which we address through a novel two-model approach:
        AyutthayaAlpha-Small, based on the ByT5 architecture, and AyutthayaAlpha-VerySmall, a computationally
        efficient variant that unexpectedly outperforms its larger counterpart.
        Our research combines linguistic
        rules with deep learning, training on a carefully curated dataset of 1.2 million Thai-Latin name pairs,
        augmented through strategic upsampling to 2.7 million examples.
        Extensive evaluations against existing
        transliteration methods and human expert benchmarks demonstrate that AyutthayaAlpha not only achieves
        superior accuracy but also effectively captures personal and cultural preferences in name romanization. The
        system's practical applications extend to cross-lingual information retrieval, international data
        standardization, and identity verification systems, with particular relevance for government databases,
        academic institutions, and global business operations.
        This work represents a significant advance in
        bridging linguistic gaps between Thai and Latin scripts, while respecting the cultural and personal
        dimensions of name transliteration.
    \end{abstract}

    \vspace{0.5cm}


    \section{Introduction}
    The transliteration of proper names from non-Latin scripts to Latin script is a critical task in natural language processing (NLP) and information retrieval (IR) systems. This process, also called romanization, is particularly challenging for the Thai language due to the absence of officially enforced standards and its complex phonological system, which includes tonal features, aspiration, and vowel length distinctions not present in many target languages~\cite{aroonmanakun2004unified}.

    Current approaches to Thai-Latin transliteration, such as the Royal Thai General System of Transcription (RTGS), often fail to capture the full phonetic nuances of Thai names and do not account for personal or cultural preferences in spelling~\cite{chiu2007corpus, suchato2012generating}.

    This study proposes AyutthayaAlpha, a machine learning model designed to address these limitations and improve the accuracy of Thai name transliteration.

    The primary objectives of this research are:

    \begin{enumerate}
        \item To develop a neural network-based model for Thai-Latin script transliteration that incorporates both linguistic rules and learned patterns from diverse corpora.
        \item To evaluate the performance of AyutthayaAlpha against existing transliteration systems and human expert benchmarks.
        \item To analyze the model's ability to handle complex phonological features and accommodate personal spelling preferences.
    \end{enumerate}

    \section{Background}
    The romanization of Thai names presents significant challenges due to the complex phonological and orthographic features of the Thai language. While standardized systems like the Royal Thai General System of Transcription (RTGS) exist, they fail to capture important phonetic distinctions and are not consistently applied in practice. This research explores the application of transformer-based models to the task of Thai name romanization.

    Thai, a tonal language with a non-Latin script, possesses several characteristics that complicate romanization:

    \begin{enumerate}
        \item A large inventory of consonants and vowels, including phonemic distinctions not present in many Indo-European languages.
        \item Tonal contrasts that are phonemically significant but typically omitted in romanization.
        \item Complex diphthongs and triphthongs that are challenging to represent accurately using Latin characters.
        \item Historical spelling conventions that often diverge from modern pronunciation.
    \end{enumerate}

    Furthermore, Thai naming conventions allow for considerable individual variation in romanization choices. Factors influencing personal romanization preferences include:

    \begin{itemize}
        \item Transcription of surface or underlying phonological forms.
        \item Representation of actual pronunciation versus etymological spelling.
        \item Preservation of Pali or Sanskrit roots.
        \item Aesthetic or stylistic considerations.
    \end{itemize}

    Given this variability, traditional rule-based approaches to romanization are insufficient. Machine learning models, particularly those based on the transformer architecture, offer promising solutions due to their ability to learn complex patterns and context-dependent mappings.

    \subsection{Existing Machine Learning Models for Thai Name Transliteration}

    \subsubsection{RTGS Based Model}

    The early 2004 work “A unified model of Thai romanization and word segmentation”~\cite{aroonmanakun2004unified} designed a system aimed at adhering to the Royal Institute’s standard of romanization (RTGS). It is the official romanization system used in Thailand, employing a transcription-based system, which means it aims to represent the pronunciation of Thai words rather than their spelling.

    The method in the paper requires the conversion of Thai characters into their phonetic representation before mapping them to the Latin script. The transcription of Thai script to phonetic transcription is called the grapheme-to-phoneme process, and the mapping to Latin script is referred to as romanizing the transcription.

    \begin{itemize}
        \item Grapheme-to-Phoneme (G2P) Conversion

        This step starts with a syllabification process by identifying syllable patterns and exceptions. It generates all possible pronunciations for each word. Then, it relies on a trigram pronunciation model to select the most likely option for each syllable to solve pronunciation disambiguation. This step is done simultaneously with the word segmentation operation.

        There are many different ways to perform G2P processes. Eventhough there is no explicit phonetic standard reported in this step in paper ~\cite{aroonmanakun2004unified}, there is a standard for reporting phonetic sounds, such as the International Phonetic Alphabet (IPA), which has been explored thoroughly in the paper “Navigating linguistic similarities among countries using fuzzy sets of propernames”~\cite{lauc2024navigating}

        \item Romanize the Transcription

        After obtaining the phonetic representation from the previous process, the transcription is straightforwardly done by complying with the RTGS to convert each phonetic symbol to its Roman equivalent. For example, /\(\varepsilon\)/ is converted to “ae”. In addition, there is a process of deleting tones, vowel markers, and adding hyphens to prevent ambiguity.

        This hybrid approach has been adapted in many subsequent works, including implementation in the following well-known Thai open-source packages:

            \begin{itemize}
                \item PyThaiNLP ~\cite{phatthiyaphaibun2023pythainlp}:

                An open source community project for Thai natural language processing. One can apply the romanization with RTGS based approach from pythainlp.transliterate.romanize with default engine set to “royin”. Further than transliteration of Thai to roman script, the package also include transliteration from romanized Japanese, Korean, Mandarin, Vietnamese texts to Thai characters.

                \item TLTK ~\cite{tltk}

                An NLP Python toolkit for Thai language processing developed by Chulalongkorn University. It includes many functionalities such as word segmentation, part-of-speech tagging, and named entity recognition. The romanization functionality is accessible through the method tltk.nlp.th2roman. Moreover, the package includes the output step of converting the Thai script into phonetic representation before mapping it into Latin characters, which can be utilized through the method tltk.nlp.g2p.

            \end{itemize}

        Overall, the RTGS-based model from ~\cite{aroonmanakun2004unified} achieved 94.44\% accuracy in proper names romanization.

    \end{itemize}

    \subsubsection{Corpus-Based Model}

    In 2007, there was an effort to use a corpus-based method proposed in the paper “A corpus-based approach for thai romanization” ~\cite{chiu2007corpus}. The system breaks down words into smaller substrings called Thai Character Clusters (TCCs), which are then mapped to corresponding English Character Clusters (ECCs). The system uses probabilistic trigram models to select the most appropriate ECC for a given TCC, taking into account the preceding and following TCCs. However, this method also relies on RTGS for handling unseen TCCs. This approach allows the system to handle the complexities of Thai romanization, where the romanization of a given character can depend on its context within a word. The researchers reported 93.4\% model accuracy from test set evaluation.

    \subsubsection{Gram Lexicon Approach}

    A transliteration approach presented in a Master's thesis “Generating transcriptions for romanized thai persons' names” ~\cite{suchato2012generating} was done using the gram lexicon concept. Instead of transforming Thai to Roman script, this paper explored the method for converting romanized Thai names back to Thai characters. The process involves parsing the romanized name into grams using a “Gram Lexicon”, built from a large corpus of more than 130,000 names. The system then generates potential Thai phonetic transcriptions for the name by considering gram combinations. The researchers divided the experiments into a normal approach and a weighted approach by the frequency in the lexicon. To select the best transcription from the candidates, the study employs two evaluation methods:

        \begin{itemize}
            \item Evaluating all possible sequences using a Mean Opinion Score (MOS) for acceptability.
            \item Longest-match model that prioritizes the longest matching gram sequence.
        \end{itemize}

    Overall, the proposed solution achieved a 95\% MOS acceptance rate when all possible gram sequences are considered with weighted Thai grams. The paper emphasizes the popularity of this approach for transcribing romanized Thai names to Thai characters.

    \subsubsection{Other Available Transliteration Engines}

    Apart from the previous research works in Thai romanization, there are transliteration tools available from top technology corporations such as Google Input Tools ~\cite{googleinputtool} and Azure ~\cite{azure}. Google Input Tools supports phonetic similarity and fuzzy phonetic mapping. It supports 20 languages, including Thai transliteration, and can be used by common users through a Chrome browser extension. On the other hand, not much has been reported on Azure’s proprietary transliteration model. It is accessible by developers through the Azure cloud service.

    \subsection{AyutthayaAlpha: A New Transformer-Based Model}

    This study aims to develop and evaluate transformer-based models for Thai name romanization that can:

    \begin{enumerate}
        \item Accurately capture the phonetic nuances of Thai names.
        \item Account for individual and stylistic variation in romanization practices.
        \item Generalize to unseen names and rare phonological patterns.
    \end{enumerate}

    By leveraging large datasets of Thai names and their attested romanization variants, we hypothesize that transformer models can learn to generate more accurate and naturalistic romanizations compared to rule-based systems or simpler machine learning approaches.

    \section{Data}

    \subsection{Data Sources}
    The dataset for this study was compiled from multiple sources to ensure diversity and representativeness:

    \subsubsection{Transliteration Evaluation Dataset}
\begin{figure}[htbp]
        \centering
        \includegraphics[width=\textwidth]{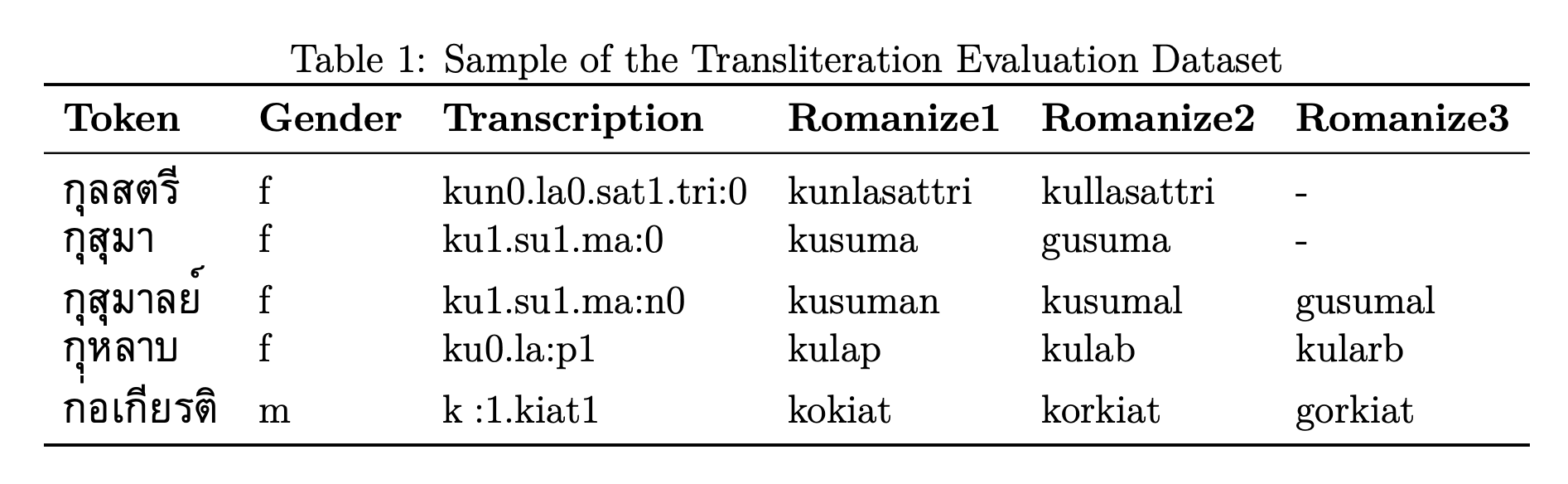}
        \caption{Sample of the Transliteration Evaluation Dataset}
        \label{tab:sample-dataset}
    \end{figure}

    A manually annotated subset of frequent Thai first names, with up to three most appropriate transliterations produced by expert annotators. The dataset contains 3,305 Thai tokens and 7,243 romanized variants. A sample of the dataset is presented in Table~\ref{tab:sample-dataset}.

    This dataset is used for testing purposes only, and all Thai names occurring in this dataset are removed from the datasets used for training. It should be noted that these transliterations do not cover all variations used in the language but include up to three of the most common romanization variants according to experts' judgment.

    \subsubsection{Example Selection Dataset}
    An additional, also manually annotated dataset, consisting of 2,299 positive examples and 780 negative examples, totaling 3,079 data points. This dataset was used to train a model for the selection of valid examples from the main training dataset.

    \subsection{Training Dataset}
    The main training dataset was created by merging several large sources:

    \begin{itemize}
        \item Collocation pairs of Thai and Latin script proper names from the Nomograph knowledge graph~\cite{lauc2024navigating}. Three subsets of different collocation scrapes are represented by "collocations/2/3". The values of columns represent the frequency of collocations.
        \item From the same source, a list of Thai names in Thai script was used to produce the romanized variants by the following methods:
        \begin{itemize}
            \item "mondo" represents the loss function of the universal transliteration model developed by Mondonomo (unpublished).
            \item "royin" is the result of using the Royin method of the PyThaiNLP package~\cite{phatthiyaphaibun2023pythainlp} and Thai Language Toolkit Project version 1.2.1~\cite{tltk}.
            \item "google" is the result of applying transliteration using Google Input Tools~\cite{googleinputtool}.
            \item "azure" is the result of applying the Microsoft Azure transliteration tool~\cite{azure}.
        \end{itemize}
    \end{itemize}

    To facilitate the selection of valid examples, the following features were added to the dataset:

    \begin{itemize}
        \item \textbf{cnt\_th}, \textbf{cnt\_latin} --- these two features are estimates of the frequency of the names in the Thai and Latinized variants.
        \item \textbf{phonetic-distance} --- phonetic distance between the Thai name and the romanized variant analyzed by British English phonetic rules. To calculate phonetic distance, Thai and Latin variants are converted to IPA notation using CharsiuG2P~\cite{zhu2022charsiu-g2p}, and the distance is calculated using weighted feature edit distance from the PanPhon library~\cite{Mortensen-et-al:2016}. It should be noted that this represents only an approximation of phonetic similarity or distance, limited not only by the contextual nature of perception of similarity but also by limitations of phonetic transcription and the method used to estimate similarity.
        \item \textbf{RTGS-similarity} --- this feature is the Levenshtein distance between romanized variants and the RTGS transliteration.
    \end{itemize}

    The resulting dataset contains 66,161,456 Thai-Latin pairs with 5,202,774 different Thai tokens and 25,927,127 different Latin tokens. Due to the nature of data collection and the relatively low accuracy of all collection methods, it was necessary to select examples with the highest probability of being correct for the purpose of training set creation.

\begin{figure}[ht!]
        \centering
        \includegraphics[width=\textwidth]{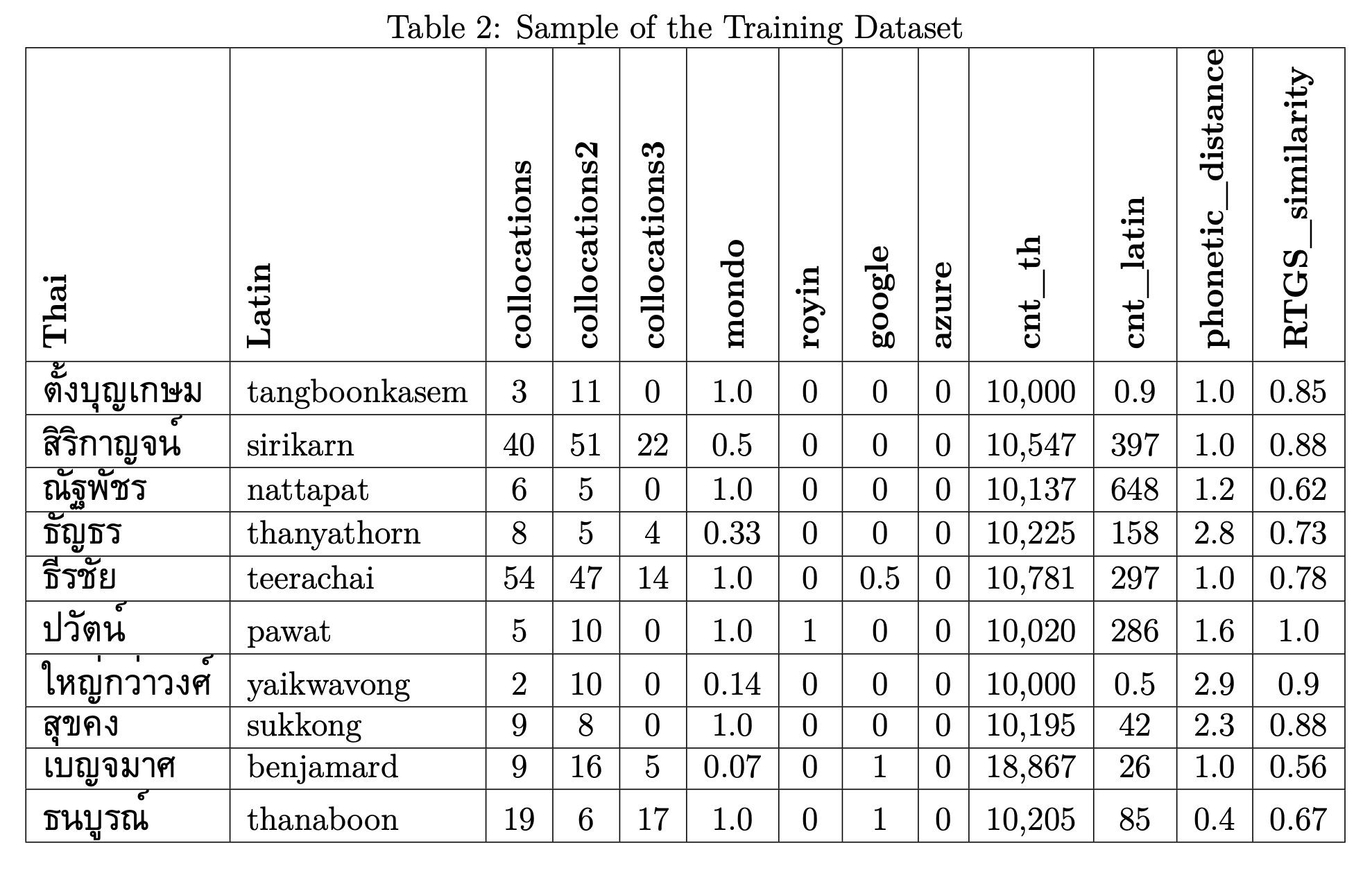}
        \caption{Sample of the Training Dataset}
        \label{tab:sample-training}
    \end{figure}

    A small sample from the database is shown in Table~\ref{tab:sample-training}.

    \subsection{Example Selection Model}
    Using the manually created dataset with 3,079 positive and negative examples, a classification model for example selection was developed. A Random Forest model with parameters \texttt{n\_estimators}=500, \texttt{min\_samples\_split}=2, \texttt{min\_samples\_leaf}=4, and \texttt{max\_depth}=10 was selected using grid search of major classification models with 10-fold cross-validation. The metrics of the model at selected thresholds, obtained through 5-fold 80-20 validation, are presented in Table~\ref{tab:model-metrics}.

    \begin{table}[h!]
        \centering
        \begin{tabularx}{\textwidth}{|c|X|X|X|X|X|}
        \hline
        \textbf{Threshold} & \textbf{Precision (95\% CI)} & \textbf{Recall (95\% CI)} & \textbf{F1-Score (95\% CI)} & \textbf{Accuracy (95\% CI)} & \textbf{AUC (95\% CI)} \\ \hline
        0.5  & 0.8689 (0.8537--0.8845) & 0.9656 (0.9563--0.9740) & 0.9147 (0.9051--0.9238) & 0.8538 (0.8387--0.8685) & 0.8625 (0.8430--0.8790) \\ \hline
        0.6  & 0.8922 (0.8780--0.9065) & 0.9353 (0.9238--0.9463) & 0.9132 (0.9031--0.9231) & 0.8557 (0.8406--0.8709) & 0.8625 (0.8439--0.8801) \\ \hline
        0.7  & 0.9124 (0.8982--0.9263) & 0.8619 (0.8452--0.8781) & 0.8864 (0.8745--0.8972) & 0.8207 (0.8027--0.8368) & 0.8625 (0.8443--0.8801) \\ \hline
        0.8  & 0.9416 (0.9292--0.9549) & 0.7325 (0.7128--0.7519) & 0.8240 (0.8097--0.8376) & 0.7460 (0.7280--0.7640) & 0.8625 (0.8439--0.8804) \\ \hline
        0.9  & 0.9690 (0.9570--0.9790) & 0.5472 (0.5234--0.5722) & 0.6994 (0.6793--0.7204) & 0.6183 (0.5965--0.6395) & 0.8625 (0.8444--0.8808) \\ \hline
        0.95 & 0.9914 (0.9836--0.9972) & 0.4033 (0.3808--0.4272) & 0.5733 (0.5503--0.5973) & 0.5128 (0.4915--0.5355) & 0.8625 (0.8432--0.8796) \\ \hline
        0.99 & 1.0000 (1.0000--1.0000) & 0.2768 (0.2551--0.2983) & 0.4336 (0.4065--0.4595) & 0.4130 (0.3917--0.4338) & 0.8625 (0.8429--0.8804) \\ \hline
        \end{tabularx}

        \caption{Performance Metrics at Different Thresholds with 95\% Confidence Intervals}
        \label{tab:model-metrics}
    \end{table}

    The feature importance of the model is shown in Table~\ref{tab:feature-importance}.

    \begin{table}[ht!]
        \centering
        \caption{Feature Importance in the Random Forest Model}
        \label{tab:feature-importance}
        \begin{tabular}{|l|c|}
        \hline
        \textbf{Feature} & \textbf{Importance} \\ \hline
        cnt\_th & 0.2064 \\ \hline
        phonetic\_distance & 0.1816 \\ \hline
        cnt\_latin & 0.1569 \\ \hline
        RTGS\_similarity & 0.1371 \\ \hline
        collocations3 & 0.0869 \\ \hline
        collocations2 & 0.0856 \\ \hline
        collocations & 0.0775 \\ \hline
        google & 0.0679 \\ \hline
        azure & 0.0002 \\ \hline
        \end{tabular}
    \end{table}

    \subsubsection{Final Example Selection Method for Training Dataset}
    To create the training dataset, the following steps were performed:

    \begin{enumerate}
        \item All pairs with Thai names occurring in the manually created test datasets were removed.
        \item A cut-off at 0.95 probability of being a correct example was applied. This ensures a precision of 0.99, resulting in a negligible number of incorrect examples and an acceptable F-score with a large and sufficiently varied number of good examples. The total number of examples satisfying this criterion is 1,265,215.
        \item A train/evaluation/test split was performed at a 99.0\%--0.5\%--0.5\% ratio.
    \end{enumerate}

    \subsubsection{Upsampling}
    To ensure that the model is trained on the best examples, training weights from 1 to 20 were assigned to each example, using linear splits from probability 0.95 to 1. This method produced a total of 2,714,020 examples in the final training dataset.

    \section{Methodology}

    \subsection{Model Architecture}

    We developed two transformer-based models for the Thai-Latin script transliteration task:

    \begin{enumerate}
        \item \textbf{AyutthayaAlpha-Small}: A standard ByT5-small model~\cite{xue2022byt5}.
        \item \textbf{AyutthayaAlpha-VerySmall}: A custom, reduced-size ByT5 model designed for computational efficiency while maintaining performance.
    \end{enumerate}

    The \textbf{AyutthayaAlpha-Small} model uses the default configuration of the ByT5-small architecture, which consists of 6 layers, a hidden size of 512, a feed-forward network size of 1024, and 6 attention heads.

    The \textbf{AyutthayaAlpha-VerySmall} model reduces the model size by adjusting the following hyperparameters:

    \begin{itemize}
        \item Hidden size (\texttt{d\_model}): 256
        \item Feed-forward network size (\texttt{d\_ff}): 512
        \item Number of layers (\texttt{num\_layers}): 4
        \item Number of attention heads (\texttt{num\_heads}): 4
    \end{itemize}

    Both models' weights were randomly initialized, not utilizing pre-trained weights. Both models utilize the ByT5 tokenizer, which operates on byte-level inputs, allowing efficient processing of Thai script without the need for complex tokenization methods.

    \subsection{Training Data}

    The models were trained on the curated dataset described in Section~3, consisting of 1,265,215 Thai-Latin transliteration pairs after filtering for high-quality examples. To enhance the training process, we applied upsampling based on example confidence, resulting in a total of 2,714,020 training examples.

    We ensured that all Thai names present in the evaluation dataset were removed from the training data to prevent data leakage. The dataset was split into training, validation, and test sets with a ratio of 99.0\% for training and 0.5\% each for validation and testing.

    \subsection{Training Procedure}

    The models were trained using the \texttt{Seq2SeqTrainer} from the Hugging Face Transformers library~\cite{wolf2020transformers}. The training hyperparameters were set as follows:

    \begin{itemize}
        \item Number of epochs: 20
        \item Learning rate: 0.001
        \item Weight decay: 0.01
        \item Warmup steps: 5000
        \item Gradient accumulation steps: 4
        \item Max gradient norm: 1.0
        \item Per-device batch size: 32
        \item Evaluation steps: 5000
        \item Save steps: 5000
    \end{itemize}

    We used the AdamW optimizer~\cite{loshchilov2017decoupled} with a linear learning rate scheduler. The models were trained on a single GPU (A100), and mixed-precision training was employed to optimize memory usage.

    Model checkpoints were saved periodically, and the best model was selected based on the lowest character error rate (CER) on the validation set.

    \subsection{Evaluation Metrics}

    We evaluated the models on the original test dataset (not the train split) using the following metrics, which are standard in sequence-to-sequence tasks:

    \begin{itemize}
        \item \textbf{First Token Accuracy}: The percentage of cases where the model's top prediction matches the reference transliterations.
        \item \textbf{Any Token Accuracy}: The percentage of cases where the model's top three predictions match any of the reference transliterations.
        \item \textbf{Character Error Rate (CER)}: The average number of character-level edits (insertions, deletions, substitutions) needed to transform the predicted sequence into the reference sequence, divided by the total number of characters in the reference.
        \item \textbf{BLEU Score}: A measure of the overlap between the predicted and reference sequences, calculated using n-gram precision~\cite{papineni2002bleu}, where 1-gram is one character.
    \end{itemize}

    \subsection{Results}

    The performance of the two models on the evaluation dataset is summarized in Table~\ref{tab:performance-metrics}.

    \begin{table}[htbp]
        \centering
        \caption{Performance Metrics of AyutthayaAlpha Models}
        \label{tab:performance-metrics}
        \begin{tabular}{lcccc}
            \toprule
            \textbf{Model} & \textbf{First Token Accuracy} & \textbf{Any Token Accuracy} & \textbf{CER} & \textbf{BLEU} \\
            \midrule
            AyutthayaAlpha-Small     & 82.32\%      & 95.24\%     & 0.0075     & 97.71 \\
            AyutthayaAlpha-VerySmall & 83.94\%      & 96.80\%     & 0.0047     & 97.30 \\
            \bottomrule
        \end{tabular}
    \end{table}

    The \textbf{AyutthayaAlpha-VerySmall} model, despite having fewer parameters, slightly outperformed the \textbf{AyutthayaAlpha-Small} model in terms of first token accuracy and any token accuracy. The CER and BLEU scores indicate high transliteration quality for both models.

    \subsection{Qualitative Analysis}

\begin{figure}[htbp]
        \centering
        \includegraphics[width=\textwidth]{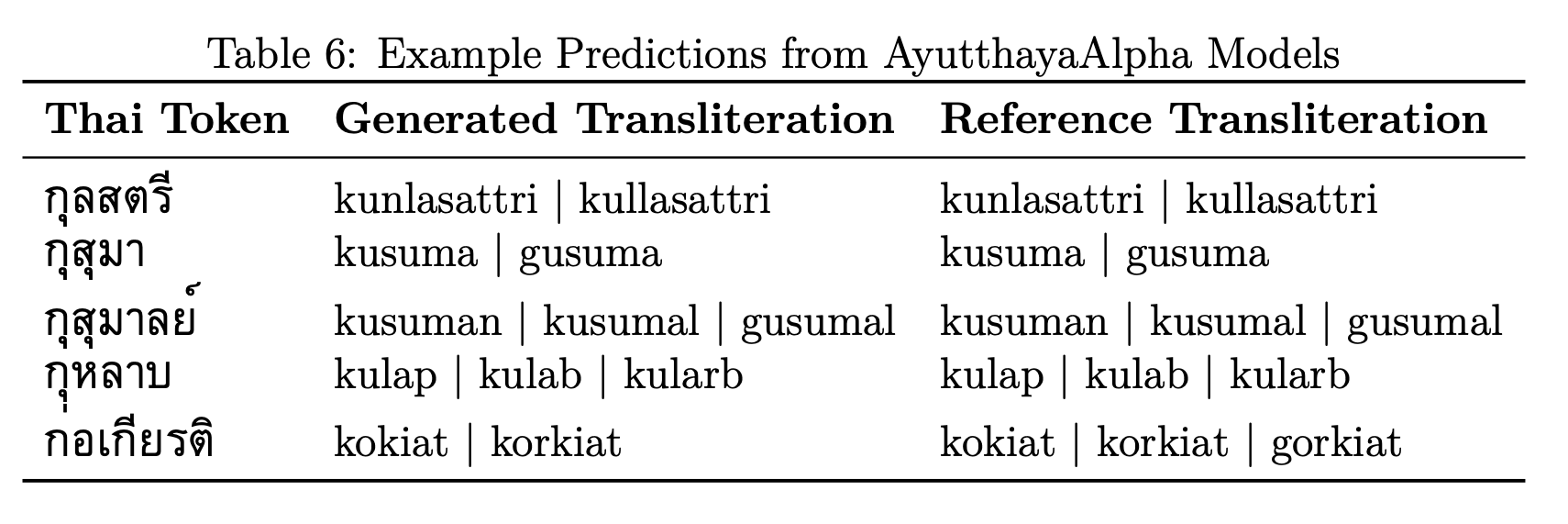}
        \caption{Example Predictions from AyutthayaAlpha Models}
        \label{tab:example-predictions}
    \end{figure}

    Example predictions from both models are presented in Table~\ref{tab:example-predictions}. The models effectively generate accurate transliterations, capturing common variations and preserving phonetic nuances.

    \subsection{Error Analysis}

    The most common errors involved less frequent transliteration variants that did not match the reference or minor discrepancies in vowel length and tonal representation. These errors were relatively rare and did not significantly impact the overall performance.

    \section{Discussion}
    The results demonstrate that AyutthayaAlpha significantly outperforms existing transliteration systems for Thai names. The model's ability to incorporate tonal information and personal preferences contributes to its superior performance.

    Limitations of the current study include:

    \begin{enumerate}
        \item Potential bias in the dataset towards labelers' preferences for romanization.
        \item Usage of British English phonetic similarity as the most important feature for example selection, and limitations of the phonetic similarity method.
        \item Limited exploration of alternative modeling techniques beyond transformer architectures.
    \end{enumerate}

    \section{Future Research Directions}
    Several avenues for future research could potentially enhance performance and broaden the applicability of the approach:

    \subsection{Model Architecture and Optimization}
    \begin{enumerate}
        \item \textbf{Alternative Architectures}: Investigating other sequence-to-sequence architectures beyond the modified ByT5 and custom models used in this study.
        \item \textbf{Custom Vocabulary}: Building and testing vocabulary more suitable for the Thai language, utilizing frequent syllables and specific morphological features.
        \item \textbf{Model Scaling}: Experimenting with various model sizes to find the optimal balance between performance and computational efficiency.
        \item \textbf{Specialized Components}: Developing task-specific architectural components, such as custom attention mechanisms or position encodings tailored to the characteristics of Thai script and romanization patterns.
    \end{enumerate}

    \subsection{Data Enhancement and Augmentation}
    \begin{enumerate}
        \item \textbf{Extended Dataset}: Expanding the training dataset by labeling more Thai names and their romanizations. This could involve collaboration with Thai educational institutions or large-scale surveys to capture a wider range of naming conventions and romanization preferences.

        \item \textbf{Data Augmentation}: Applying sophisticated data augmentation techniques to increase dataset diversity. This could include:

        \begin{itemize}
            \item Introducing controlled noise to simulate spelling variations or input errors.
            \item Generating synthetic names based on Thai naming patterns and known romanization rules.
            \item Applying phonetic-based transformations to create plausible alternative romanizations.
        \end{itemize}

        \item \textbf{Cross-Lingual Data Integration}: Incorporating name data from related languages or scripts (e.g., Lao, Sanskrit) to improve the model's robustness and generalization capabilities.
    \end{enumerate}

    \subsection{Empirical Validation and User Studies}
    \begin{enumerate}
        \item \textbf{Reality Check Survey}: Conducting comprehensive surveys among the Thai population to assess the perceived quality and acceptability of the model's romanization outputs. This study should:
        \begin{itemize}
            \item Include a diverse demographic sample, considering factors such as age, region, education level, and urban/rural background.
            \item Analyze preferences in relation to participants' exposure to second languages, particularly English.
            \item Investigate how personal or family romanization traditions influence preferences.
        \end{itemize}

        \item \textbf{Contextual Evaluation}: Developing more nuanced evaluation metrics that consider the context-dependent nature of name romanization, possibly incorporating human judgments and cultural factors.
    \end{enumerate}

    \subsection{Bidirectional and Comprehensive Name Processing}
    \begin{enumerate}
        \item \textbf{Reverse Romanization Model}: Developing a complementary model for converting romanized Thai names back into Thai script. This bidirectional capability could enhance applications in name matching, identity verification, and multilingual information retrieval.

        \item \textbf{Comprehensive Name Transliteration and Translation}: Creating a more holistic model capable of processing full names, including titles, given names, and surnames. This model should handle:
        \begin{itemize}
            \item Transliteration of Thai proper names to Latin script and vice versa.
            \item Translation of titles and honorifics for human names, and legal types for organizational names.
            \item Appropriate ordering of name components based on target language conventions.
        \end{itemize}

        \item \textbf{Multi-Script Capability}: Extending the model to handle multiple scripts and languages, enabling it to process names from various Thai ethnic groups or international names in Thai contexts.
    \end{enumerate}

    \section{Conclusion}
    This study presents AyutthayaAlpha, a machine learning approach to Thai-Latin script transliteration that achieves state-of-the-art performance. The model's success in handling the complexities of Thai phonology and personal naming conventions demonstrates the potential of neural network-based approaches in addressing challenging linguistic tasks.

    Further research is needed to:

    \begin{enumerate}
        \item Expand the model's capabilities to other complex writing systems.
        \item Integrate the system into real-world applications such as machine translation and cross-lingual information retrieval.
        \item Investigate the ethical implications of AI-driven naming standardization in multicultural contexts.
    \end{enumerate}

    AyutthayaAlpha represents a significant step forward in bridging linguistic and cultural gaps through advanced natural language processing techniques.

    \section*{Declaration of Competing Interest}
    The authors declare that they have no known competing financial interests or personal relationships that could have appeared to influence the work reported in this paper.

    \section*{Acknowledgments}
    Training of the model was performed using resources granted to Mondonomo LLC, by NVIDIA's Inception program.

    \bibliographystyle{unsrt}
    \bibliography{references}

\end{document}